# System theoretic approach of information processing in nested cellular automata

Jerzy Szynka

Retired researcher at the Łukasiewicz Research Network Institute of Microelectronics and Photonics (formerly the Institute of Electron Technology), Warsaw, Poland

**Abstract**

The subject of this paper is the evolution of the concept of information processing in regular structures based on multi-level processing in nested cellular automata. The essence of the proposed model is a discrete space-time containing nested orthogonal space-times at its points.
The factorization of the function describing the global behavior of a system is the key element of the mathematical description. Factorization describes the relations of physical connections, signal propagation times and signal processing to global behavior.
In the model appear expressions similar to expressions used in the Special Relativity Theory.

## 1. Introduction

The idea of information processing in regular structures was first put forward by John von Neumann, Stanislaw Ulam, and Konrad Zuse. This idea was later developed in various aspects under the following names: cellular automata, computing space, tessellation structure, systolic array, threshold network [1],[2],[3]. Today, this concept and its implementation in silicon is being realized under the name of neuromorphic computing e.g. Intel Neuromorphic Research Community.

Research into information processing (signal and image recognition, data mining, deep learning) in regular structures was inspired by neural networks and is being currently applied to artificial intelligence solutions.

A popular cellular automaton model is a regular network of interconnected cells performing identical logical functions.

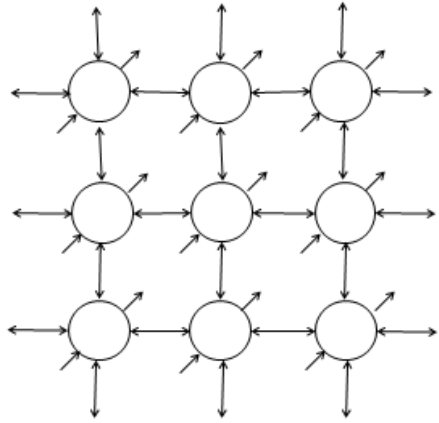

Fig.1    2-dimensional cellular automaton.

This paper starts with a theoretical description of a cellular automaton, and then extends this description to discuss multi-nested cellular automata model whose characteristic features are nested orthogonal space-times. Algebraic description for such model will be presented.

## 2.    Mathematical description of nested cellular automaton

The starting point is a cellular automaton. For the clarity of description and illustration, the description of the cellular automaton is presented for one-dimensional space $R$.

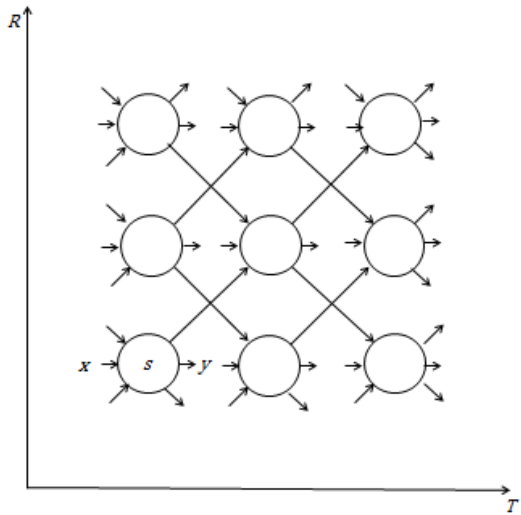

Fig.2    Propagation and processing of signals in the cellular automaton.
       $R$ is discrete space, $T$ is discrete time, $x$ is the input, $s$ the state and $y$ the output of a cell.

Globally, the processing of signals in discrete space-time $R \times T$ is described by the function
$A: S^{R \times T} \times X^{R \times T} \to S^{R \times T} \times Y^{R \times T}$, where $S^{R \times T}$ is the set of mappings of space-time $R \times T$ into a set of states $S$, (distributions of states in the space-time). Similarly $X^{RxT}$ is the set of input distributions and $Y^{RxT}$ is the set of output distributions.

The global function $A$ can be presented as a factorization of two functions $B$ and $C$ related to the functions defining the cellular automaton [4],[5].

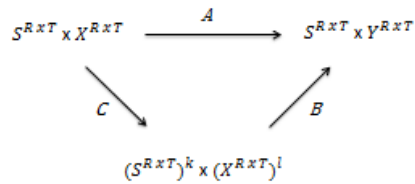

Fig.3    Factorization diagram of global function of cellular automaton.

The local behavior of a cellular automaton is defined by:

space-time structure       $v: R \times T \rightarrow (R \times T)^{k+l}$        $v(r, t) = ((r_1, t_1), ..., (r_{k+l}, t_{k+l}))$ and the cell function $F \times G: S^k \times X^l \rightarrow S^k \times Y^l$        $s^k = F(s_1,...,s_k,x_1,...,x_l)$   $y^l = G(s_1,...,s_k,x_1,...,x_l)$

These two functions are related to both factorials $B$ and $C$ of the global function $A$ in the following way. The $v$ function determines shifts of the distributions $S^{RxT}$ and $X^{RxT}$ such that $k$- and $l$-factors differ only in shifts in time and space. The $F$ and $G$ functions operate on these multiples pointwise for every point $(r, t) \in R \times T$. The result is a diagram describing the overall cellular automaton.

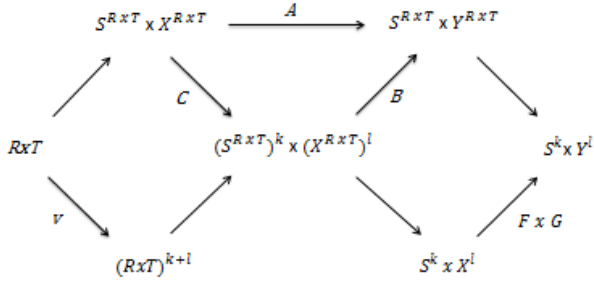

Fig.4 The relationship between the global function $A$ of a cellular automaton factorized by $C$ and $D$ and the local cell functions which describe the space-time structure $v$ and the cells local functions - state transition $F$ and output function $G$.

The local functions have the form:

$s^k(r,t) = F(s(r-r_1, t-t_1), …, s(r-r_k, t-t_k), x(r-r_1, t-t_1), …, x(r-r_l, t-t_l))$

$y^l(r,t) = G(s(r-r_1, t-t_1), …, s(r-r_k, t-t_k), x(r-r_1, t-t_1), …, x(r-r_l, t-t_l))$

where the argument shift defines the space-time structure from which the propagation speeds of the signals $v_1 = r_1/t_1$, $v_2 = r_2/t_2$ … can be calculated.

In the case of a multidimensional space, further propagation directions occur.

## 2.1    Nested space-time

In nature, each object is composed of sub-objects whose interaction determines the action of the whole. This (two level) local-global relationship is the key research problem.

The objective of this paper is to extend this consideration to cover multiple levels of nesting and their mathematical description.

Turning to the mathematical description, the important observation is that the local function $F \times G$: $S^k \times X^l \to S^k \times Y^l$ shows a formal similarity with $A: S^{R \times T} \times X^{R \times T} \to S^{R \times T} \times Y^{R \times T}$. Consequently, the multiplicities $k$ and $l$ can be interpreted as the internally nested space-time $R' \times T'$ of the cell.

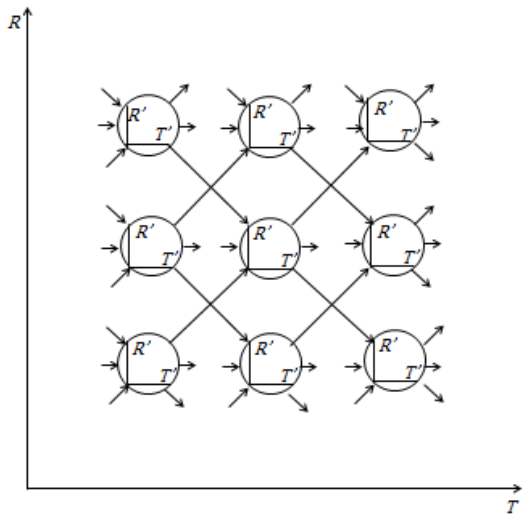

Fig.5   Nested cellular automaton.

Such situation describes the extended factorization diagram.

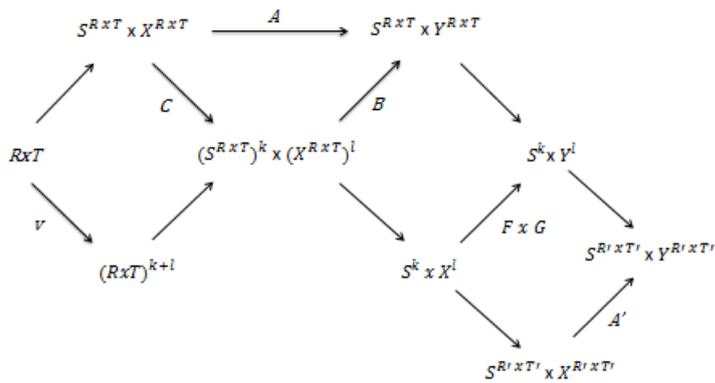

Fig.6   Factorization of global function of nested cellular automaton.

As a continuation of this idea, further levels of nested space-times can be created.

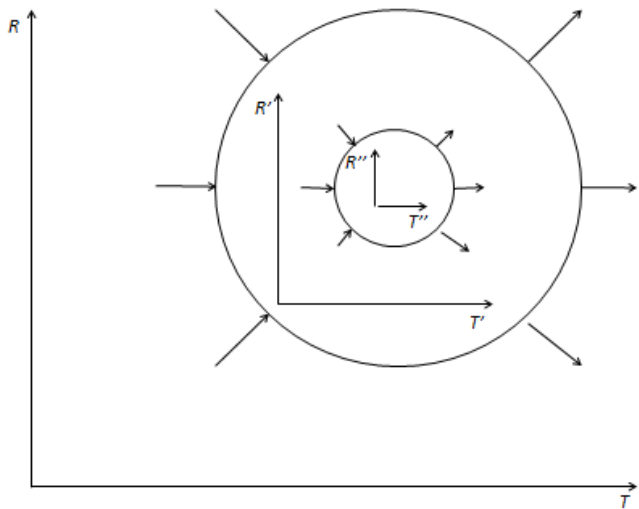

Fig. 7   Double nested cellular automaton.

At each point in space-time $R \, x \, T$ there is a nested space-time $R' \, x \, T'$, and at each point of the space-time $R' \, x \, T'$, there is a nested space-time $R'' \, x \, T''$.

Let us note that the signal processing functions occur only at the lowest level of nesting.
This is consistent with the situation in neural networks, where the neurons are surrounded by a large number of branching connections and realize a relatively simple threshold function themselves.

**2.2   Signal propagation and processing in orthogonal space-times**

Assuming that the nested space-times are orthogonal to higher level space-times, at each point in the space-time $R \, x \, T$ there is a nested space-time $R' x T'$, which is orthogonal to $R \, x \, T$.

Because the space-times are orthogonal, the propagation speed vectors are also orthogonal.

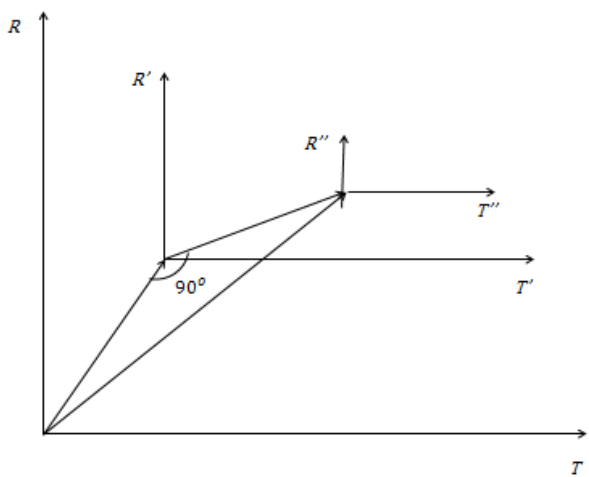

Fig.8   Signal propagation in nested space-times.

The three orthogonal space-times need in fact six coordinates. Therefore, when looking at the figure providing a two-dimensional plane, one must remember that they are properly orthogonal. The apparent parallelism of the axes of space and time in the drawing is due only to limited possibilities of presenting the situation as a figure.

The propagation speed in nested space-time can be calculated as follows:

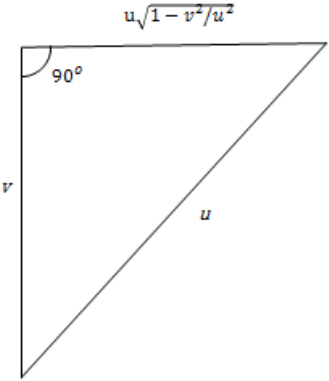

Fig.9   Calculation of speed in nested space-time.

It follows from the right triangle formula that, based on two sides of the triangle, the third side can be calculated using the expression $u\sqrt{1 - v^2/u^2}$.

When the symbol for the (global) speed $u$ is replaced by the symbol for the speed of light $c$, we obtain an expression that appears in the Special Relativity Theory $c\sqrt{1 - v^2/c^2}$. The question arises if this is a coincidence, or a fact of wider significance, indicating that the relativistic factor is related to the orthogonality of space-times.

In the case of double nesting where each point of space-time $R \times T$ contains at its points an orthogonal space-time $R' \times T'$ and it in turn contains at each point orthogonal space-time $R'' \times T''$, the sum of the relative speeds $v, w$ of the individual space-times appears in the expression of the magnitude of the third side.

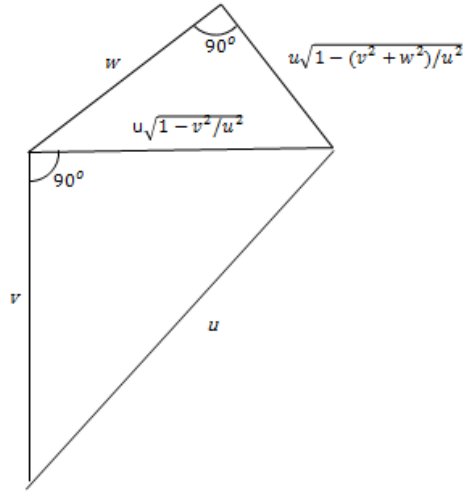

Fig.10  Calculation of speed in double-nested space-time.

Based on the above considerations, an expression for the propagation speed of signals in space-times with different levels of nesting can be formulated as follows:

$r_1/t_1 = v_1, ..., r_k/t_k = v_k$ (the zero level of nesting)

$r'_1/t'_1 = u\sqrt{1 - v_1^2/u^2}, ..., r'_k/t'_k = u\sqrt{1 - v_k^2/u^2}$ (the first level of nesting)

$r''_1/t''_1 = u\sqrt{(1 - (v_1^2 + w_1^2)/u^2)}, ..., r''_k/t''_k = u\sqrt{1 - (v_k^2 + w_k^2)/u^2}$ (the second level of nesting)

The following is the general propagation and signal processing formula for space-time up to the second level of nesting:

$s((r,t), (r',t'),(r'',t'')) = F(s((r - r_1, t - t_1), ..., (r - r_k, t - t_k), (r'-r'_1, t'-t'_1), ..., (r' - r'_k, t' - t'_k), (r''-r''_1, t''-t''_1), ..., (r'' - r''_k, t'' - t''_k)))$

which, for special cases, takes this form:

a) Signal propagation without processing in 2-level space-time:

A case analogous to that occurring in the Special Relativity Theory.

$s((r,t), (r',t')) = s((r - r_1, t - t_1), (r'- r'_1, t'-t'_1))$

$= s((r - v_1 t_1, t - t_1), (r' - u\sqrt{1 - v_1^2/u^2}\, t'_1, t' - t'_1))$

b) Signal propagation with processing in 3-level space-time:

$s((r,t), (r',t'),(r'',t'')) = F(s((r - r_1, t - t_1), (r'-r'_1, t'-t'_1), (r''-r''_1, t''-t''_1)))$

$= F(s((r - v_1 t_1, t - t_1), (r' - u\sqrt{1 - v_1^2/u^2}\, t'_1, t' - t'_1), (r'' - u\sqrt{(1 - (v_1^2 + w_1^2)/u^2)}\, t''_1, t''-t''_1)))$

c)    Propagation of multiple signals of different speeds and processing in 2-level space-time:

$$s((r,t),(r',t'),(r'',t'')) = F(s((r-r_1, t-t_1),\ldots, s(r-r_k, t-t_k), (r'-r'_1, t'-t'_1),\ldots, (r'-r'_k, t'-t'_k)))$$

$$= F(s((r-v_1 t_1, t-t_1),\ldots, s((r-v_k t_k, t-t_k), (r'-u\sqrt{1-v_1^2/u^2}\, t'_1, t'-t'_1),\ldots, (r'-u\sqrt{1-v_k^2/u^2}\, t'_k, t'-t'_k)))$$

Similarly, signal processing formulas can be formulated for more complex models with multidimensional space (different propagation directions) which are multi-level and include inputs and outputs.

## 3.    Concluding remarks

The presented concept of information processing in nested regular structures is attractive because of its similarities to the most popular form of information presentation - the textual form. The basic structure of textual information is characteristically multi-level. A word is a sequence of letters. A sentence is a sequence of words. A paragraph is a sequence of sentences, and so on. In the language of algebra, where $S$ represents the basic set of letters, words are a subset of $S^R$, sentences are a subset of $(S^R)^{R'}$, and paragraphs are a subset of $((S^R)^{R'})^{R''}$. Thus, we have a textual information structure relevant to the structure of the proposed model.

A question arises regarding the technical implementation of this concept. There are three basic options:

- with software, relying on existing computers.

- with software and hardware, relying on current technologies. This implementation is currently realized in the framework of the Intel Neuromorphic Research Community initiative.

- hardware implementation in materials.

And just as integration in silicon represented a breakthrough in the transition of electronics to microelectronics, a hardware implementation of the concept discussed in this paper holds promise.